\documentclass[letterpaper, 10 pt, journal, twoside]{sty/IEEEtran}
\usepackage{graphicx}
\usepackage{amsmath}
\usepackage{multirow}
\usepackage{comment}

\begin{document}
\title{
    Deep Active Visual Attention for Real-time Robot Motion Generation:
    Emergence of Tool-body Assimilation and Adaptive Tool-use
}

%

\author{
    Hyogo Hiruma$^{1}$, Hiroshi Ito$^{1,2}$, Hiroki Mori$^{3}$, and Tetsuya Ogata$^{1,4}$
    \thanks{
        Manuscript received: February, 24, 2022; Revised May, 27, 2022; Accepted June, 22,
        2022. This paper was recommended for publication by Editor A. Banerjee upon
        evaluation of the Associate Editor and Reviewers' comments. This work was supported
        by Hitachi, Ltd.
    }
    \thanks{
        $^{1}$Hyogo Hiruma, Hiroshi Ito and Tetsuya Ogata are with the Department of
        Intermedia Art and Science, Waseda University, Tokyo, Japan
        {\tt\footnotesize hiruma@idr.ias.sci.waseda.ac.jp, hiroshi.ito.ws@hitachi.com, ogata@waseda.jp}
    }%
    \thanks{
        $^{2}$Hiroshi Ito is with the Center for Technology Innovation - Controls and Robotics,
        Research \& Development Group, Hitachi, Ltd., Ibaraki, Japan
    }%
    \thanks{
        $^{3}$Hiroki Mori is with the Future Robotics Organization, Waseda University, Tokyo, Japan
        {\tt\footnotesize mori@idr.ias.sci.waseda.ac.jp}
    }%
    \thanks{
        $^{4}$ Tetsuya Ogata is with the Waseda Research Institute for Science and
       Engineering (WISE) at Waseda University, and the National Institute of
       Advanced Industrial Science and Technology (AIST), Tokyo, Japan
    }%
    \thanks{Digital Object Identifier (DOI): see top of this page.}
}

\markboth{
    IEEE Robotics and Automation Letters. Preprint Version. Accepted June, 2022
}{
    Hiruma \MakeLowercase{\textit{et al.}}: Deep Active Visual Attention for Real-time Robot Motion Generation
} 

\maketitle

\begin{abstract}
Sufficiently perceiving the environment is a critical factor in robot
motion generation. Although the introduction of deep visual processing models
have contributed in extending this ability, existing methods lack in the ability
to actively modify what to perceive; humans perform internally during
visual cognitive processes. This paper addresses the issue by proposing
a novel robot motion generation model, inspired by a human cognitive structure.
The model incorporates a state-driven active top-down visual attention module,
which acquires attentions that can actively change targets based on task states.
We term such attentions as role-based attentions, since the acquired attention
directed to targets that shared a coherent role throughout the motion.
The model was trained on a robot tool-use task, in which the role-based attentions
perceived the robot grippers and tool as identical end-effectors, during object
picking and object dragging motions respectively. This is analogous to a biological
phenomenon called tool-body assimilation, in which one regards a handled tool
as an extension of one's body. The results suggested an improvement of flexibility
in model's visual perception, which sustained stable attention and motion even
if it was provided with untrained tools or exposed to experimenter's distractions.
\end{abstract}

\begin{IEEEkeywords}
Neurorobotics, Bioinspired robot learning, Visual Attention Mechanism, Imitation learning
\end{IEEEkeywords}

%
\IEEEpeerreviewmaketitle

\section{INTRODUCTION}
\IEEEPARstart{U}{nderstanding}
the relationship between the body and environment is an essential
task for both robots and humans, in order to accurately predict the results
of a motion. Humans associate the relationships, or body schema, through body
babbling and dynamic touches \cite{gibson}, both of which infants perform to
empirically verify the link between vision, haptics, and self produced
motions.

In robot task learning, conventional rule-based methods used manually
pre-programmed relationships which generate precise motions, but were
only applicable to strictly controlled environments. Conversely,
learning-based methods acquire relationships by learning through interactions
of trial and error \cite{reinf_levine, reinf_gu}, imitating demonstrated
motions \cite{imitation_yokoya, imitation_uchibe}, or motor babbling
\cite{takahashi}, which is inspired by human's body babbling behaviours.
In particular, end-to-end learning methods
\cite{takahashi, itosan, ichiwarasan} directly associated visual and
motor sensory data as body schema, which induced vision modules to selectively
extract image data that are task-relevant. However, the environment modeled by
such vision modules tended to be biased by the training data, which failed to
adapt to recognize environment at untrained situations. Since the construction
of body schemata strongly depends on visual information, such a lack of
adaptability induces distorted environment recognition.

\begin{figure}[t]
\centering
  \includegraphics[width=0.88\linewidth]{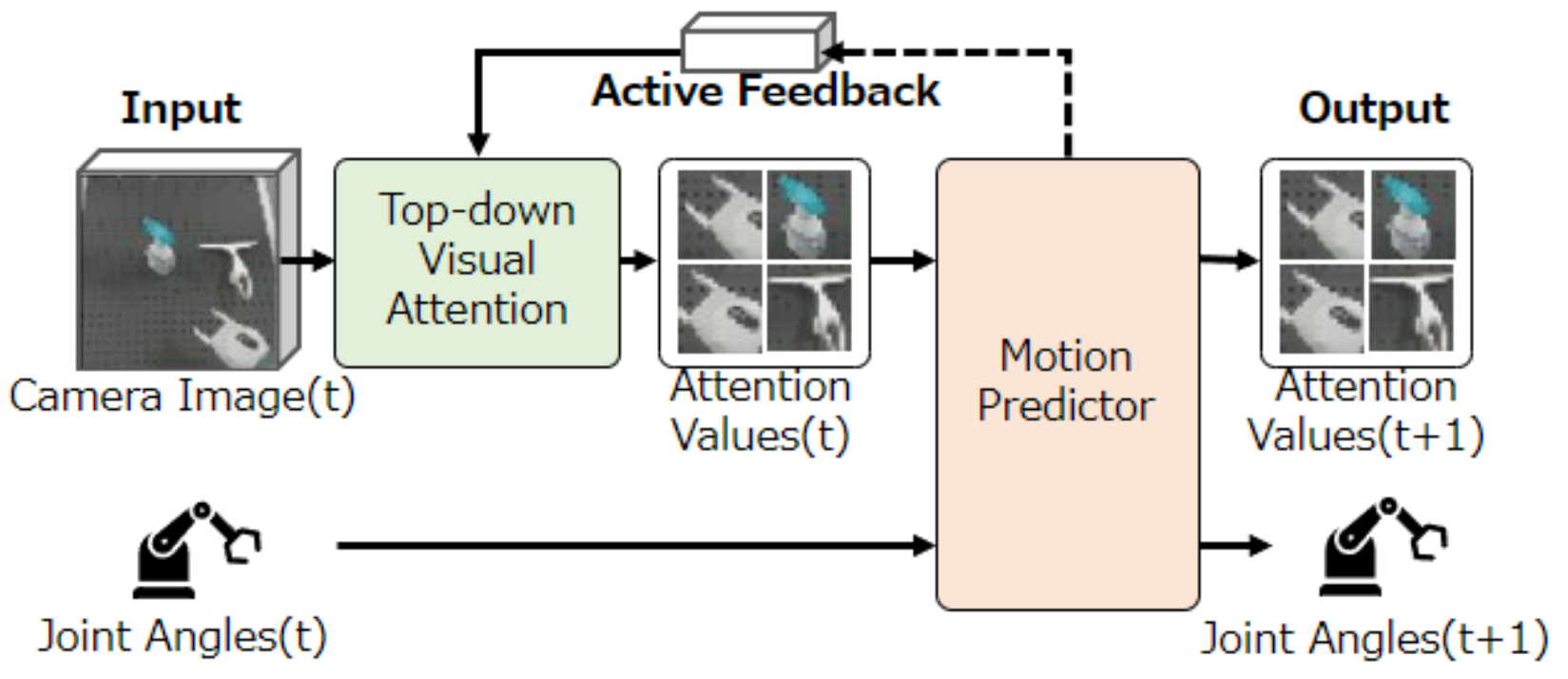}
  \caption{
    Abstract structure of the proposed motion generation model. The top-down 
    visual attention module extracts informative visual data from camera
    image based on active feedback of the succeeding motion predictor module.
  }
  \label{fig:abstract_structure}
 \vspace{-5mm}
\end{figure}

The visual perception of humans utilizes a cognitive structure called visual attention
\cite{desimone_duncan, infant_attention}, which selectively extracts information
from attended local visual area. Human visual attention is an integrated system
of bottom-up (passive) attention and top-down (active) attention. The former
passively collects conspicuous visual patterns, whereas the latter actively filters
out the collected stimuli to acquire the desired information. The desired information
are selected based on what is held in one's working memory. Since the active attention
enables self-control of the visual perception targets, humans are capable of attending
to appropriate targets based on different situations. We consider that such ``activeness'',
which existing robot vision models lack, enables humans to efficiently model environments
with less biased recognition.

\begin{figure*}[t]
\centering
  \includegraphics[width=0.85\linewidth]{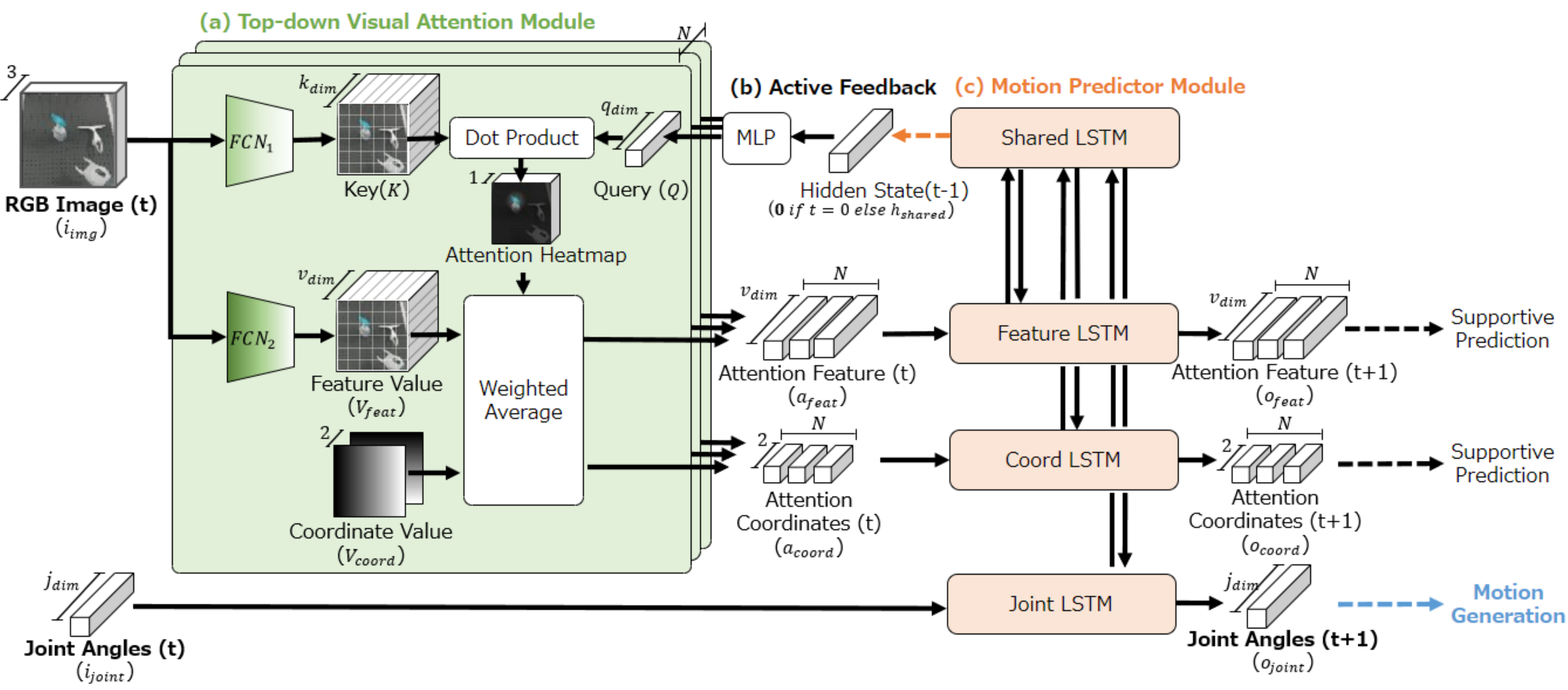}
  \caption{
    Detailed structure of the proposed model: (a) Top-down visual attention
    module, (b) Active feedback connection and (c) Motion predictor module.
    The attention module selects and extracts feature and coordinate values at multiple visual
    attention points and the motion module uses them along with current robot joint angle values,
    for predicting each future values. Via active feedback, the attention targets are constantly
    updated based on the internal states of the motion module.
  }
  \label{fig:model_structure}
  \vspace{-5mm}
\end{figure*}

This paper proposes a novel real-time robot motion generation model, as shown in 
Fig. \ref{fig:abstract_structure}, inspired by human cognitive structures. The model
is composed of a top-down attention module and a motion predictor module, in which
the internal states of the latter module are constantly fed back to actively modify
attention targets. The model was trained on a robot tool use task, which acquired
role-based attentions in contrast to conventional appearance-based attentions.
The attention indicated visual recognition similar to Japanese macaques, suggesting
an emergence of a biological phenomenon called tool-body assimilation.
\section{RELATED WORK}

\subsection{Visual perception in robot control}
In robot tasks, camera images are commonly utilized as the main source for
perceiving the environment. However, as raw pixel data include irrelevant
noises, directly using image data leads to a distorted perception, which
negatively affects the construction of the robot's body schema.

Various vision processing models have been proposed to extract informative
data from camera images, such as rule-based saliency maps \cite{saliency_map}.
Studies have proposed learning-based vision models that solely learn object-centric
representations \cite{fasterrcnn, yolo} to explicitly extract data of certain objects.
By contrast, end-to-end learned models jointly train vision models along with robot control
models. This makes the vision models, such as widely used vanilla fully convolutional
networks (FCN), to extract task-oriented features even from untrained objects
\cite{itosan, saitosan, ichiwarasan}. However, FCNs lack in spatial generalization
ability \cite{coordconv}, and cannot extract data at untrained visual areas.
Although training with large dataset can mitigate such inability, the data collection
cost is a large burden especially; this is a common issue when employing other visual
attention models such as vision transformers \cite{vit, detr}.

Recent visual attention models specialized for robots \cite{ichiwarasan, DSAE, TransporterNet}
achieved high spatial generalizability with data efficiency, by employing a structure that
explicitly extracts coordinate values of the predicted visual attention points. However,
the inherent structure which assumes the attention targets to have consistent appearances
removed the ability to generalize to untrained objects. While some models enabled top-down
control of attention targets based on human inputs \cite{MyWork, TransporterNet2}, they cannot
perform on site self-controlling which is to direct to necessary targets based on current
motion states. The proposed model adds ``activeness'', as in humans, to self-control the
attention targets, by providing motion prediction feedback to the visual attention module.
We term our model as ``Active'' in contrast to other existing ``Passive'' attention models.

\subsection{Tool-use learning}
\label{sec:tool_use_learning}
How one perceives a tool is an important metric when evaluating how
it recognizes the relation between itself and the environment. Tool use
behaviors are exhibited by animals to overcome their physical limitations.
Animals show high dexterity by perceiving the tool as if it is a part of
their bodies \cite{iriki, hikita}. Iriki et al. \cite{iriki} suggested that
Japanese macaques perceived the tool as an extension of their body after
being trained to use them (tool-body assimilation). Specifically, such
assimilation did not occur when macaques ``held'' the tool. Rather, it
occurred when they were trained to ``use'' the tools. This suggests that
actively perceiving to associate bodies, objects, and environments is
essential to efficiently learn to use unaccustomed tools. Accordingly,
how one recognizes the tool indicates how flexibly one can redefine the body schema.

Given the flexible tool-use behaviours of humans, previous studies of
robot tool-use \cite{takahashi, nabeshima, stoytchev} proposed methods
that enable the robots to control different tools in a single model,
by predicting tool inertia parameters. The parameters were predicted
by tool-held motor babbling \cite{takahashi}, computing from the observed
force and torques when swinging tools \cite{nabeshima}, and exploratory
behaviours \cite{stoytchev}, which enabled the model to generalize to
handle untrained tools. However, as the parameters cannot be predicted
on the fly, the models cannot predict motions that requires to automatically
switch tools based on transitioning situations. This suggests the lack in
capability to redefine the body schema in real-time. Therefore, this research
focuses to verify the flexible recognition of the proposed method, by
evaluating whether it holds the capacity to redefine the body schema without
specialized training methods for tool-body assimilation.
\section{PROPOSED METHOD}

\subsection{Design concept}
\label{sec:design_concept}
The proposed model is a robot motion generation model which actively modifies
what to perceive in a task-state-driven and top-down manner (Fig.
\ref{fig:model_structure}). The inputs are image data $i_{img}^{t}$ and
robot joint angle data $i_{joint}^{t}$ at time step $t$, and the output
is the prediction of next joint angles $o_{joint}^{t+1}$. The model generates
robot motions by repeating the sequence of predicting the next joint angles
and applying them to the robot \cite{itosan,saitosan,ichiwarasan}.

The model consists of a top-down visual attention module (Fig.
\ref{fig:model_structure} (a)) and a motion predictor module (Fig.
\ref{fig:model_structure} (c)) which are connected with a forward pass and an
active feedback (Fig. \ref{fig:model_structure} (b)). The visual attention module
is based on our previous ``Passive'' top-down attention model \cite{MyWork},
but is extended to preform ``Active'' on site target control, by the whole
structure of this model. This paper distinguishes ``Active'' and ``Passive'' attention
by whether the model self-controls the attention via temporal internal feedbacks,
or is controlled via external inputs, respectively; models that do not incorporate
top-down information is also considered ``Passive'' (e.g. self-attention). The motion
predictor module is composed of a ``Layered'' Long-Short Term Memory (LSTM) structure
which is inspired by \cite{goal_directed_attention}, in contrast to existing models that
output visuomotor predictions with a ``Single'' LSTM. This structure enables higher
and lower layered LSTMs to predict transitions of abstract task states and raw
modality data respectively. Especially, separating the abstract transition is
beneficial for creating active feedbacks that reflect slowly transitioning task
states instead of rapidly changing raw modality values.

The structures of this model are inspired by human cognitive structures. The
active feedback connection of the two modules is analogous to human visual attention,
which controls attention targets based on information that are held in the working
memory. The motion predictor module plays the role of the working memory. It is also
designed to mimic the function of human working memory which simultaneously possess
both integrated (abstract) and unintegrated (raw) data of different modalities, or stimuli
\cite{wm_integration}.

\subsection{Top-down visual attention module}
This module (Fig. \ref{fig:model_structure}(a)) predicts task-relevant visual
areas and selectively extracts the corresponding feature and coordinate values
\cite{MyWork}. The module generates key-query-value representations, which is
trained to extract information that are task-relevant. The key $K$ is an image
feature of input $i_{img}$, and the query $Q$ is a vector that represents the feature of
attention targets. The value consists of two representations: the feature value
$V_{feat}$ and coordinate value $V_{coord}$. The former is an image feature of
the input $i_{img}$, and the latter is an absolute position embedding of Cartesian
coordinates. This model converts attention maps to coordinate and feature values
to effectively compress the data for the succeeding motion prediction and to perform
object spatial recognition with high data-efficiency, as described in \cite{MyWork}.

The attention is computed in two phases. The first phase generates an attention
heatmap $M$ through the scaled channel-wise dot product of $K$ and $Q$, followed
by a spatial softmax function. The heatmap places high values at pixels of $K$ that
contain features similar to vector $Q$, which indicate the existence of an attention
target. The second phase computes a weighted average of $V_{feat}$ and $V_{coord}$
in spatial dimensions by utilizing $M$ as the weight of each pixel. This
selectively extracts values at attention-target pixels: $a_{feat}$ and $a_{coord}$.

\begin{equation}
    M = Softmax_{2D}
        \left(
            \frac{ K \odot Q }{ \sqrt{HW} }
        \right)
\end{equation}
\begin{equation}
    a_{feat} = 
        \sum_{u=0}^{W}
        \sum_{v=0}^{H}
        M_{(u, v)} V_{feat (u, v)}
\end{equation}
\begin{equation}
    a_{coord} =
        \sum_{u=0}^{W}
        \sum_{v=0}^{H}
        M_{(u, v)} V_{coord (u, v)}
\end{equation}
where $H$ and $W$ represent the unified height and width, respectively, of
$K$, $M$, and $V$. $(u, v)$ represents the coordinates at each representations.
This attention module can determine the attention target by utilizing different
$Q$s, which can be generated using various methods. The proposed model utilizes
the internal states of the motion prediction module, which will be discussed later.
The attention module also can be extended to multi-head attention. The outputs of
each attention head are concatenated and denoted as $A_{feat}$ and $A_{coord}$ in
hereafter.

\subsection{Motion predictor module}
This module (Fig. \ref{fig:model_structure}(c)) integrates data of three
modalities, which are attention features $A_{feat}^{t}$, coordinates $A_{coord}^{t}$,
and joint angles $i_{joint}^{t}$, and predicts each value at the next
time step. The predicted joint angles $\hat{o}_{joint}^{t+1}$ are utilized for
motion generation, and the predicted attention values $\hat{A}_{feat}^{t+1}$ and
$\hat{A}_{coord}^{t+1}$ are utilized as supportive data for training the entire
model. The module is a layered structure composed of four units of LSTMs.
Three LSTMs are lower-layered ``modal LSTMs'', that predict future values of individual
modalities. The remaining is an upper-layered ``shared LSTM'', which integrates the
internal states of the modal LSTMs and redistributes to share the states of
different modalities. This layered structure is inspired by the characteristics
of human cognitive activity, where information on visual features, positions and
motions does not interfere within the working memory
\cite{wm_visuospatial, wm_visuomotor}. $h$ denotes the hidden states of each LSTMs.

\begin{equation}
    \left\{
        \begin{aligned}
            \hat{A}_{feat}^{t+1}, h_{feat}^{t} &=
                LSTM_{feat}(A_{feat}^{t}, h_{feat}^{t-1}, h_{shared}^{t-1}) \\
            \hat{A}_{coord}^{t+1}, h_{coord}^{t} &=
                LSTM_{coord}(A_{coord}^{t}, h_{coord}^{t-1}, h_{shared}^{t-1}) \\
            \hat{o}_{joint}^{t+1}, h_{joint}^{t} &=
                LSTM_{joint}(i_{joint}^{t}, h_{joint}^{t-1}, h_{shared}^{t-1}) \\
        \end{aligned}
    \right.
\end{equation}

\begin{equation}
    \left\{
        \begin{aligned}
            h_{shared}^{t} &=
                LSTM_{shared}(h_{modal}^{t-1}, h_{shared}^{t-1}) \\
            h_{modal}^{t-1} &= Concatenate(h_{feat}^{t-1}, h_{coord}^{t-1}, h_{joint}^{t-1})
        \end{aligned}
    \right.
\end{equation}

\begin{figure}
\centering
  \includegraphics[width=\linewidth]{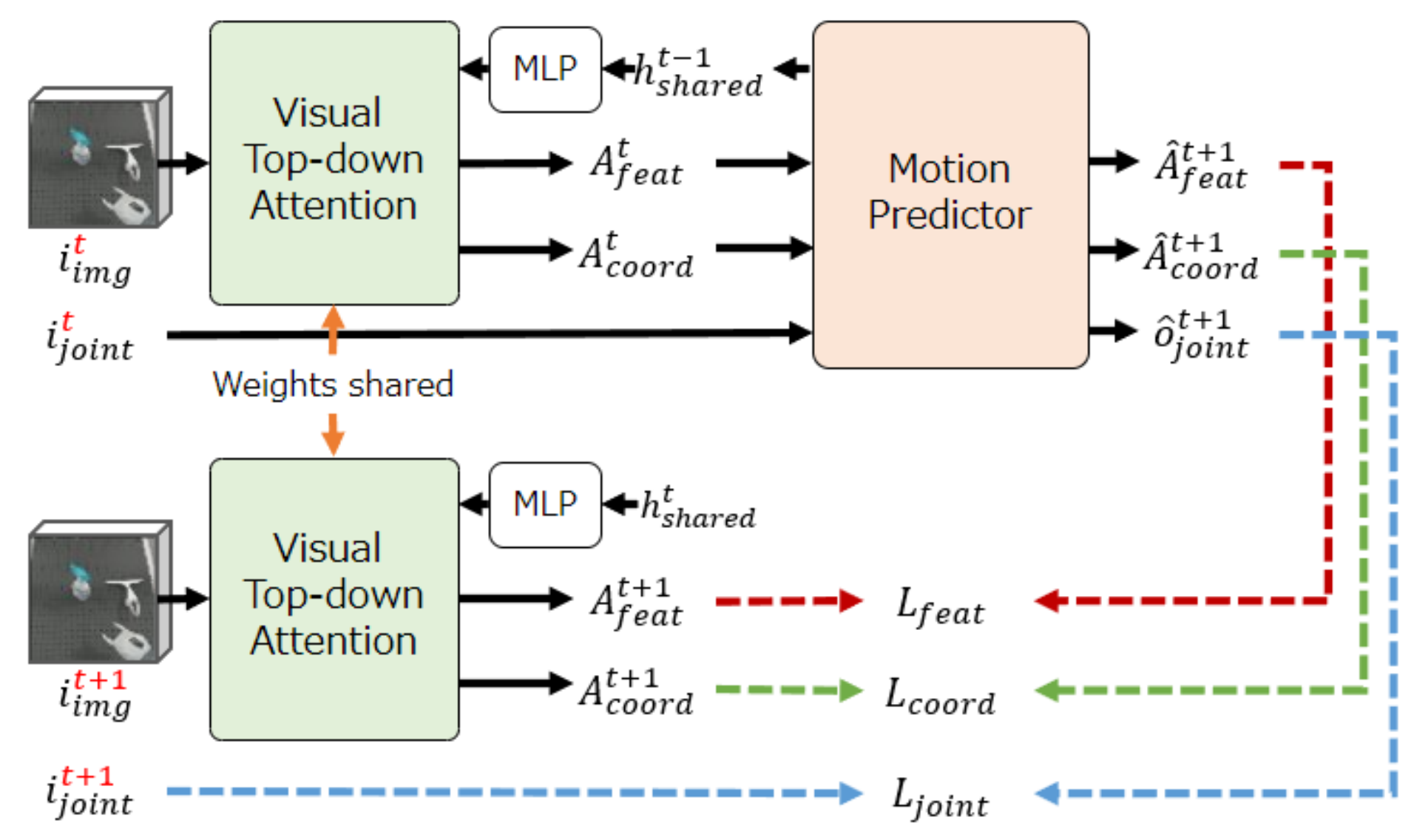}
  \caption{
    Process of bi-directional loss. The loss of attention values predicted
    by the motion predictor module is computed with those predicted by
    the attention module using actual future images. Best viewed in color.
  }
  \label{fig:bidirectional_loss}
\end{figure}

\begin{figure*}[t]
\centering
  \includegraphics[width=0.95\linewidth]{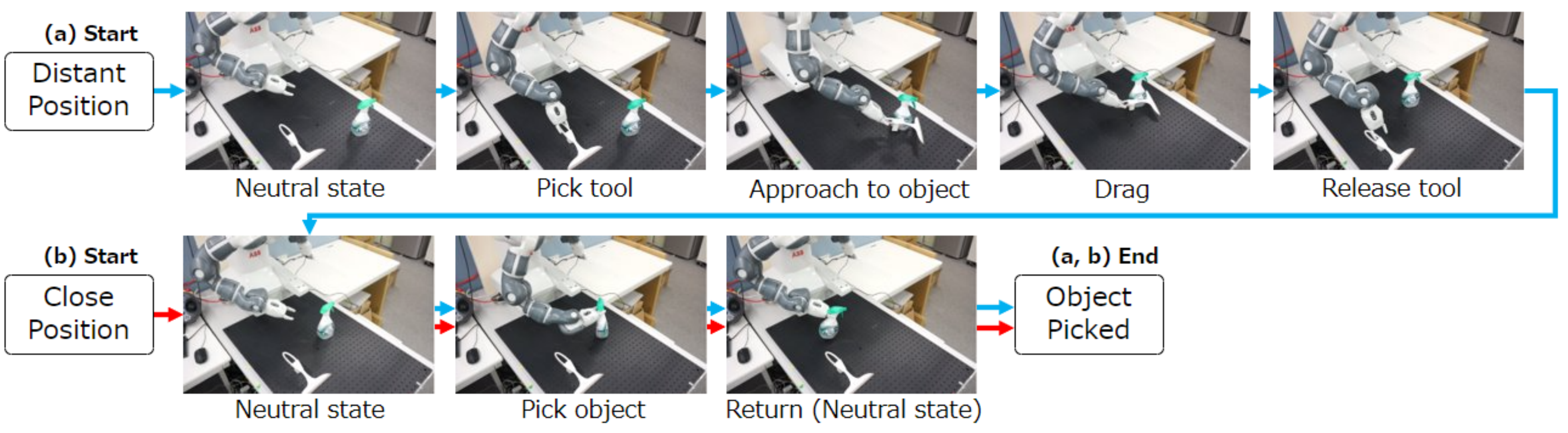}
  \caption{
    Overall task flow of the trained motions. The objective is to pick the object.
    (a) When the object is placed at a distant position, the robot drags the object
    with a tool and then picks it up with the gripper. (b) When the object is placed
    nearby, the robot directly picks up the object with the gripper.
  }
  \label{fig:experiment_task}
\end{figure*}

\subsection{Query generation based on active feedback}
In this model, the query vector for top-down attention is generated based on  
active feedback data from the motion predictor module (Fig.
\ref{fig:model_structure}(b)). Concretely, the previous internal states of the
shared LSTM $h_{shared}$ are converted to a query vector with multilayer perceptrons
(MLPs). As the shared LSTM integrates data of all modalities, we considered that
the body and the environmental information are structured within. Therefore, the
generated $Q$s are expected to acquire attention for suitable targets at each
time step by considering the situation during task execution. On the initial time
step, the $Q$ was generated from a zero vector.

\subsection{Bi-directional Loss}
The proposed model is trained by optimizing it to minimize the error $L$, which
is a weighted sum of future visuomotor prediction errors of modal LSTMs:
$L_{feat}$, $L_{coord}$, and $L_{joint}$; the applied weights are
$\alpha_{feat}$, $\alpha_{coord}$, $\alpha_{joint}$ respectively. Minimizing
the joint error ($L_{joint}$) leads to precise motion prediction, while
minimizing the attention errors ($L_{feat}$ and $L_{coord}$) leads to better
recognition of the environment. However in the proposed model, the ground
truth values of attention outputs do not exit. Therefore, this model employs
a temporally bi-directional loss function, inspired by \cite{bidirectional_loss},
which computes errors with self-simulated future values instead of ground-truth
values. As Fig. \ref{fig:bidirectional_loss} shows, future attention outputs
predicted by LSTMs $\hat{A}_{feat}^{t+1}$ and $\hat{A}_{coord}^{t+1}$, are
compared with $A_{feat}^{t+1}$ and $A_{coord}^{t+1}$. The comparison targets
are predicted by a weight shared visual attention module, which is input the
actual future image $i_{img}^{t+1}$ in the dataset. This method is memory
efficient compared to previous methods \cite{itosan, saitosan, ichiwarasan},
that compute loss of reconstructed images with actual future images. The
bi-directional loss not only eliminates redundant image reconstructions, but
also induce the visual attention and motion prediction modules to predict attention
points that are mutually consistent. In addition, this supports the layered LSTM to
structure representations for active feedback, that better consider temporal context of
visual data. Mean squared error (MSE) was used to compute the error of each prediction.

\begin{equation}
    \left\{
        \begin{aligned}
            L &= \alpha_{feat}L_{feat} + \alpha_{coord}L_{coord} + \alpha_{joint}L_{joint} \\
            L_{feat} &= MSE(\hat{A}_{feat}^{t+1}, A_{feat}^{t+1}) \\
            L_{coord} &= MSE(\hat{A}_{coord}^{t+1}, A_{coord}^{t+1}) \\
            L_{joint} &= MSE(\hat{o}_{joint}^{t+1}, i_{joint}^{t+1}) \\
        \end{aligned}
    \right.
\end{equation}
\section{EXPERIMENTS}

\subsection{Environmental setup}
\begin{figure}
\centering
  \includegraphics[width=\linewidth]{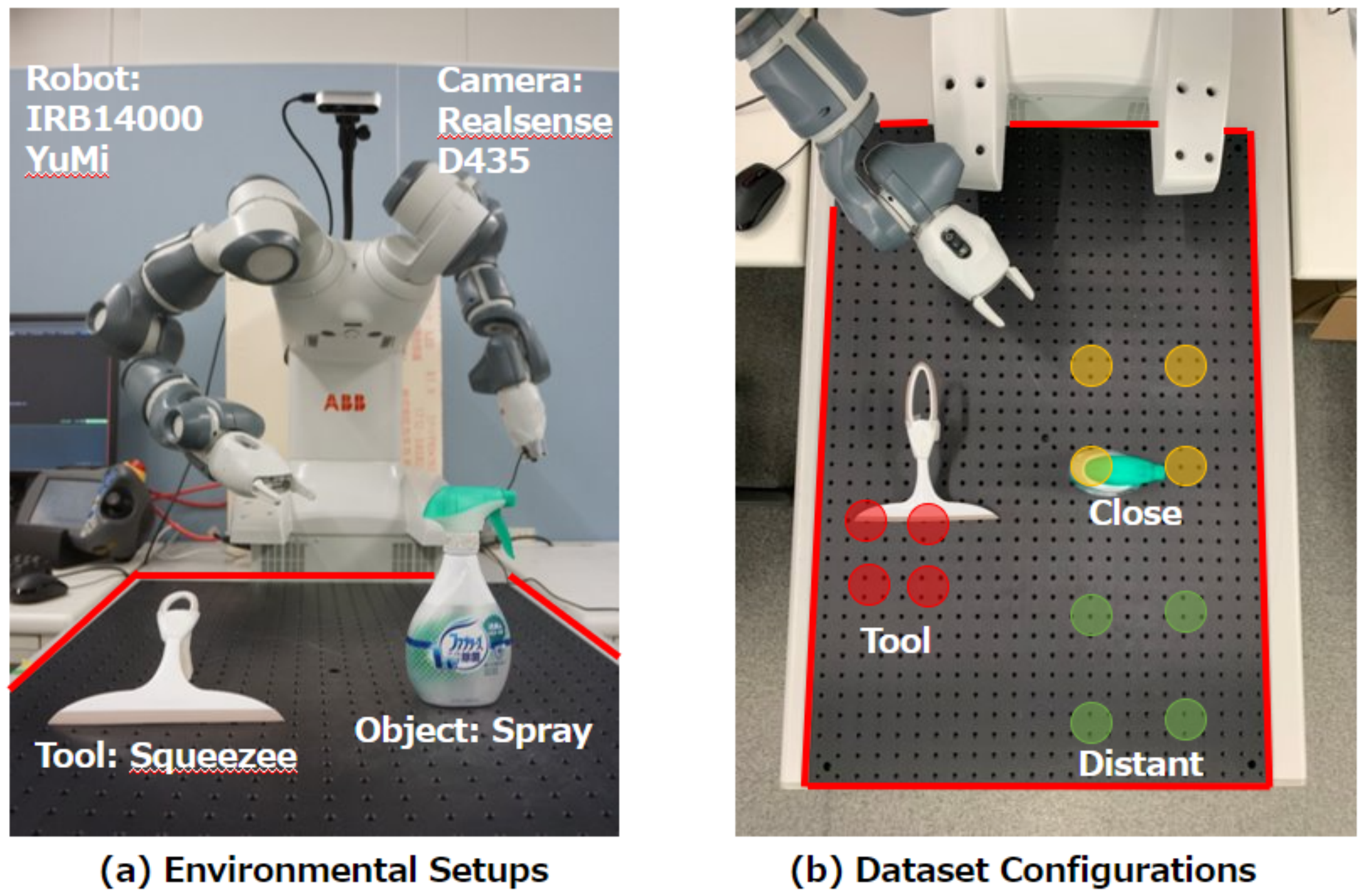}
  \caption{
    Environmental setup of the experiment. (a) Hardware, tool, and object
    viewed from the front. (b) Positions where the tool and object were placed
    in the training data. Red indicates tool positions, yellow indicates
    close object positions and green indicates distant object positions
  }
  \label{fig:experiment_setups}
 \vspace{-3mm}
\end{figure}

Fig. \ref{fig:experiment_setups}(a) displays the environmental setup of the
experiment. The robot is ABB IRB 14000 YuMi, which is a dual-arm robot with each
arm having seven degrees of freedom, and a gripper is installed. The camera on the
robot is an Intel RealSense D435 RGBD camera (depth images were not used in this
experiment). A white squeezee was used as a tool and a spray bottle as the
target object of interaction.

\begin{figure*}[t]
\centering
  \includegraphics[width=0.90\linewidth]{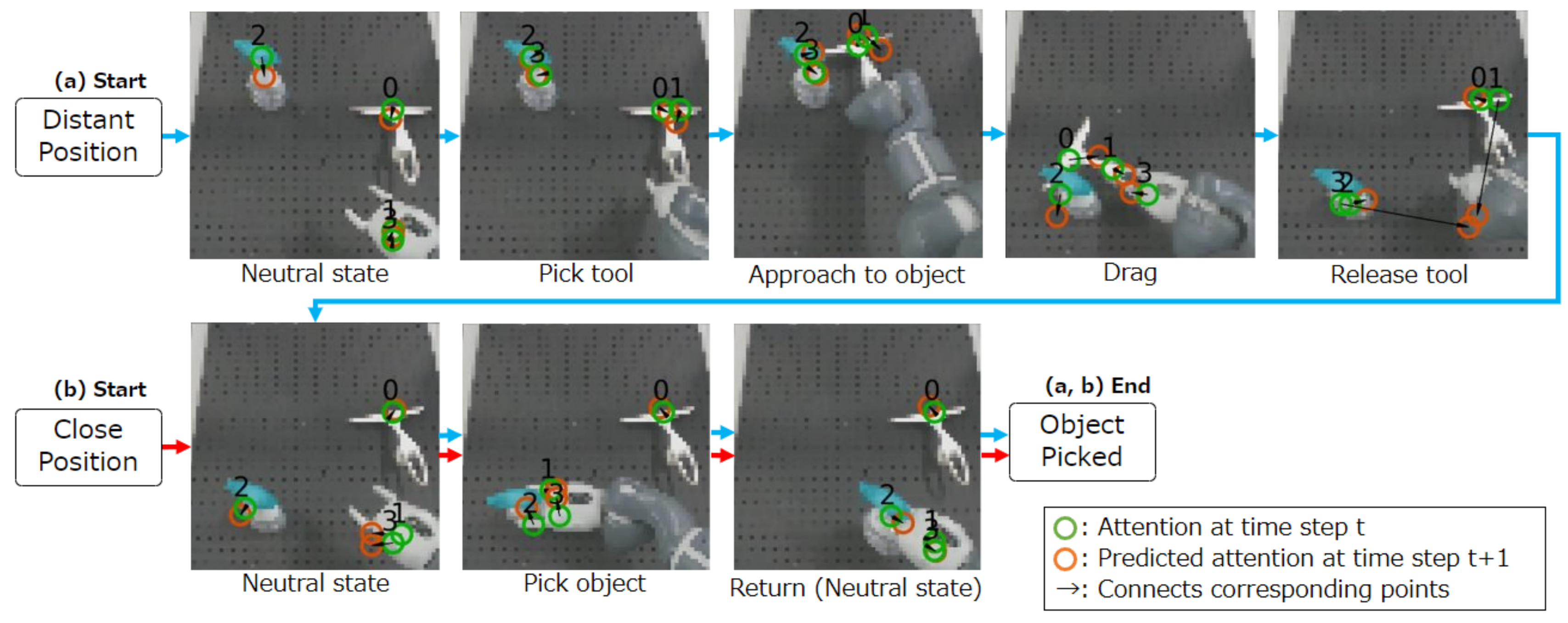}
  \caption{
    Behavior of predicted attention points during consecutive dragging-picking
    motions. Attention \#0 and \#2 are appearance-based attentions, that constantly
    acquire attention for fixed targets. Attention \#1 and \#3 are role-based
    attentions that shift between targets based on a coherent role.
  }
  \label{fig:normal_results}
  \vspace{-3mm}
\end{figure*}

\begin{table*}
    \centering
    \caption{Motion success rate under different conditions}
    \scalebox{1.0}{
        \begin{tabular}{l|c |c c|c c c} \hline
              & Proposed Model & \multicolumn{2}{c|}{Existing Models} & \multicolumn{3}{c}{Ablation Studies}\\
            \hline
            Motion Type & Active + Layered & Vanilla FCN & FCN Attention & Active + Single & Passive + Layered & Passive + Single \\
            \hline
            Picking (close object) & \textbf{27}/30 & 0/30  & 3/30  & 0/30 & 0/30  & 0/30 \\
            Dragging (distant object) & \textbf{30}/30 & 20/30 & 22/30 & 22/30 & 15/30 & 9/30 \\
            Consecutive & \textbf{26}/30 & 0/30  & 3/30  & 0/30 & 0/30  & 0/30 \\
            \hline
        \end{tabular}
    }
    \label{tab:success_rate}
    \normalsize
    \vspace{-3mm}
\end{table*}

\subsection{Task settings}
The robot tool-use task was designed to induce active changes of attention
targets. As Fig. \ref{fig:experiment_task} shows, the overall goal of the task
is to pick the spray bottle object. However, different motions are performed
depending on the position of the spray bottle. The robot directly picks the spray
bottle with a gripper if it is placed nearby, whereas the robot drags the spray
bottle with a tool before picking if it is placed at a distance. Hence, feature extraction
of different visual targets are necessary in each motion: spray bottle and gripper
for the picking motion, and spray bottle and tool data for the dragging
motion. This experiment was also designed to be analogous to the task
in which tool-body assimilation was observed in Japanese macaques \cite{iriki}.

\subsection{Training configurations}
The model is trained on a dataset which consists of sequence data of camera images and
joint angles (including gripper states) during the above-mentioned motions. Twelve
sequences of each picking, dragging, and consecutive dragging-picking motion were collected;
the length are 60, 60, and 120 timesteps, respectively. The object was placed at four different
positions of close and distant range, and the tool was placed at four different positions
as well. 

The model was not only trained to learn the motion data, but also to self-determine
which type of motion to generate, depending on the given situation. We expect that such
a training induces the model to actively modify the attention targets when transitioning
to different states. For training, the internal states of all LSTMs were
restricted to match the initial/neutral state to enable smooth transition between
motions, as in \cite{kasesan}. 

In the experiment, $FCN_{1}$ and $FCN_{2}$ were 3 layer FCNs, with
$filters=\{30, 30, q_{dim}\times N\}$, $kernel=3\times3$, and $stride=2$. The model
was set with $k_{dim}=q_{dim}=10$ and $v_{dim}=10$. The number of attention heads is $N=4$,
in reference to the average number of features a human can hold in the visual working memory.
The MLP of active feedback consists of four fully connected layers, with
$nodes=\{300, 100, 60, q_{dim}\times N\}$. The shared and modal LSTMs were
$nodes=20$ and $nodes=30$ respectively. The loss weights were $\alpha_{feat}=0.001$,
$\alpha_{coord}=0.0001$, and $\alpha_{joint}=0.9$. The model was trained by an ADAM
optimizer with learning rate of 0.001, batch size 5, for 15000 epochs. 
\section{RESULTS AND DISCUSSION}

\subsection{Predicted motion and attention}
Table \ref{tab:success_rate} presents the motion success rate of the proposed
model, which uses an ``active'' visual attention module and a ``layered'' LSTM
motion predictor module. The success rate was measured for picking, dragging, and
consecutive motions with 30 trials each, where the object and tool were
placed at untrained positions. The motions were considered to have succeeded
if the spray bottle was manipulated as expected. The results showed high precision
for all motions, and the model succeeded in precisely generating motions of
the corresponding types. In an additional experiment, the model was presented with
an unexpected situation, where the experimenter relocates the object to a distant
position immediately after the model completed the dragging motion. The model
handled this situation by regenerating the dragging motion instead of transitioning
to the picking motion, which suggests that the model successfully learned to
determine the motion based on the object location; more details are shown in the video.

Fig. \ref{fig:normal_results} shows the four predicted attention points during
consecutive dragging-picking motions, all of which stably acquired attention for 
different targets. Specifically, two among the four continuously acquired attention
for the same targets, while the other two acquired attention for different targets
during different motions. Table \ref{tab:targets_and_motions} summarizes the
relationship between the targets of each four attention heads and the type of predicting
motions. While attention \#0 and \#2 consistently directed attention to the same targets, 
attention \#1 and \#3 directed to visually dissimilar targets. However, considering
the relation with the corresponding motions, the information extracted from such targets
can be interpreted to have similar roles: the information extracted from attention \#1
can be interpreted as the end-effector data, and that from attention \#3 as supportive
data. In this regard, we consider \#1 and \#3 to have acquired consistent role-based
attention, in contrast to \#0 and \#2, which are conventional appearance-based attentions
\cite{DSAE, TransporterNet, MyWork}. The existence of role-based attention suggests that
the model actively modified the attention targets by considering the current situation
that are fed back from the shared LSTM.

\begin{table}
    \centering
    \caption{Relationship between attention targets and motions}
    \scalebox{1.0}{
        \begin{tabular}{c|c|c} \hline
             Attention Head & Neutral/Picking & Dragging  \\
             Index          & (Object: Close) & (Object: Distant) \\
            \hline
            0 & Tool & Tool \\
            1 & \textbf{Gripper} & \textbf{Tool} \\
            2 & Object & Object \\
            3 & \textbf{Gripper} & \textbf{Object} \\
            \hline
        \end{tabular}
    }
    \label{tab:targets_and_motions}
    \normalsize
 \vspace{-3mm}
\end{table}

\subsection{Comparison with existing models}
The proposed model was compared to two existing robot motion prediction models.
The models are similarly structured with two modules, but with different vision
models: (1) ``Vanilla FCN'' model \cite{itosan, saitosan}, which uses
vanilla FCNs to extract the feature of the whole image. (2) ``FCN Attention''
model, which directly predicts attention points image feature of FCN output
\cite{DSAE, ichiwarasan}. This model is ``passive'', as it does not incorporate
active feedback from a motion prediction module. The two models are trained to
minimize motion prediction errors, along with future image prediction error as
an auxiliary loss.

\begin{figure}
\centering
  \includegraphics[width=\linewidth]{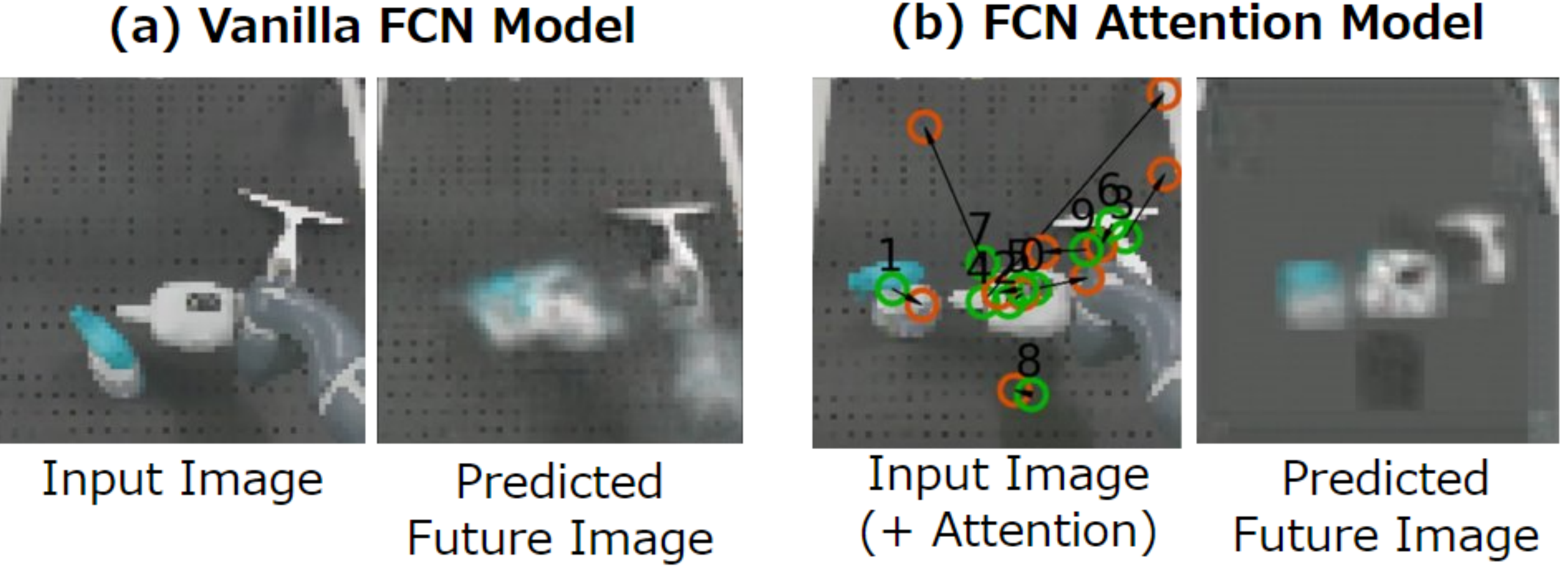}
  \caption{
    Prediction results of existing models
  }
  \label{fig:other_results}
 \vspace{-3mm}
\end{figure}

Table \ref{tab:success_rate} lists the respective task success rates and Fig.
\ref{fig:other_results} shows the predicted attention points and future images.
The success rates of the two models were low compared to the proposed model,
especially on tasks that involve object picking. This is because picking requires
higher precision compared to dragging, since there is only little difference in
width between an opened gripper and spray. Specifically, ''vanilla FCN'' model
performed poor due to the misrecognition of object positions, which can be observed
from the predicted image in Fig. \ref{fig:other_results} (a); the object was
reconstructed at an irrelevant image area. ``FCN Attention'' model succeeded
in acquiring necessary attentions (gripper, tool, and object), and to generate
appropriate motions at untaught situations. However, the model tended to perform
picking at positions where the motion prediction module expected the object to be,
rather than the actual positions perceived by the vision modules (Fig.
\ref{fig:other_results}(b)). More details are shown in the video. This is caused by
the over-dependence on the auxiliary loss, in which the motion prediction is
largely affected by the image prediction result. Such dependence also reduces the
interpretability of the predicted future attention points, as shown in Fig.
\ref{fig:other_results}.

\subsection{Ablation study on proposed modules}
To verify the functions of the two modules of the proposed model, an ablation study
was conducted. Each module was modified and trained with all combinations of the
following conditions: (1) use ``Active'' or ``Passive'' attention module and (2) use
``Layered'' or ``Single'' LSTM module; each conditions are explained in
Section \ref{sec:design_concept}. Among the four combinations, ``Active + Layered''
is the originally proposed structure. The modified models were trained and evaluated
using the same dataset. As summarized in Table \ref{tab:success_rate}, the performance
substantially  decreased with all modified structures. The following describes the
predicted behavior of motions and attentions.

\begin{figure*}[t]
\centering
  \includegraphics[width=0.9\linewidth]{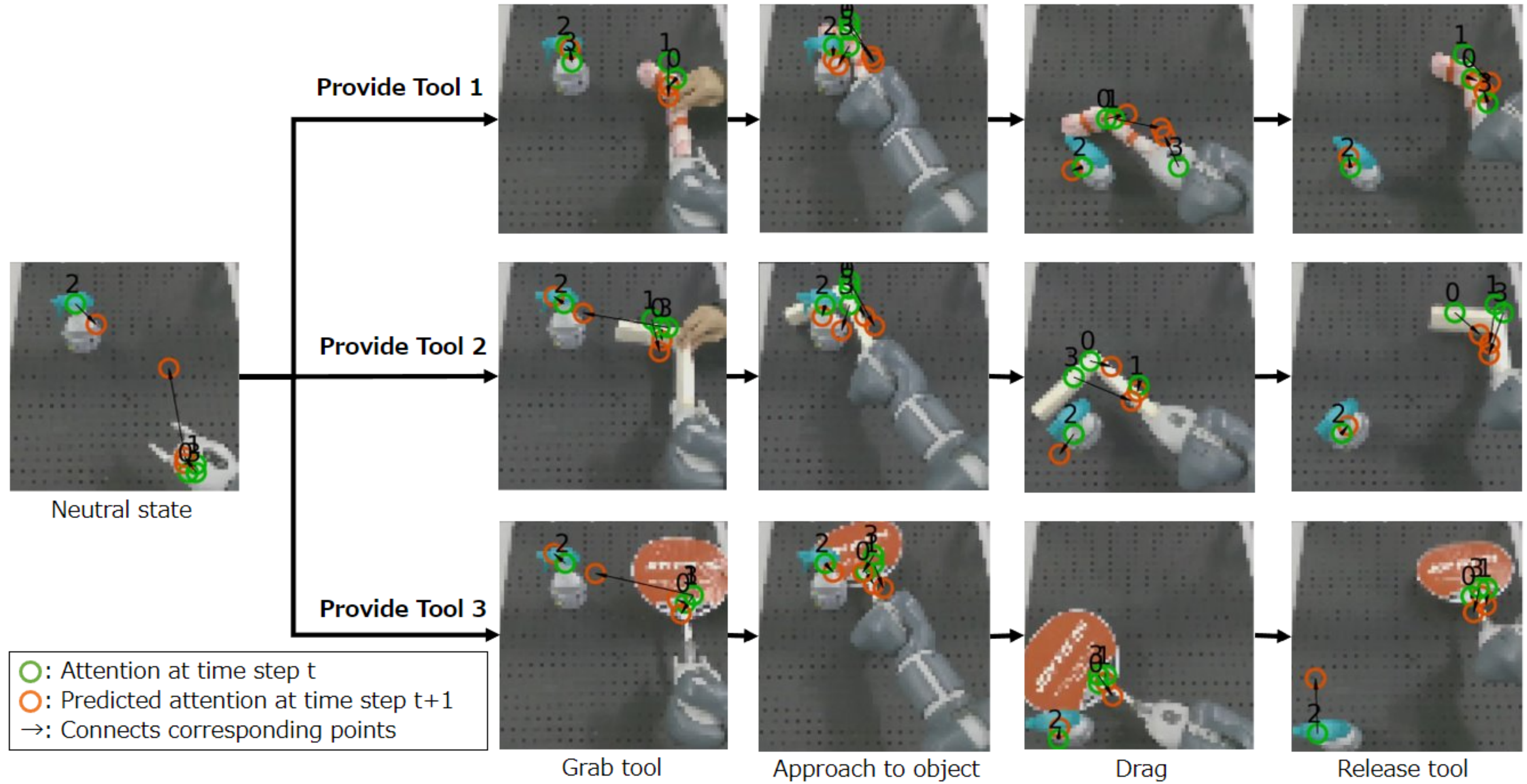}
  \caption{
    Predicted motion and attention when provided with untrained types of tools.
  }
  \label{fig:tool_results}
 \vspace{-3mm}
\end{figure*}

\begin{figure*}[t]
\centering
  \includegraphics[width=0.9\linewidth]{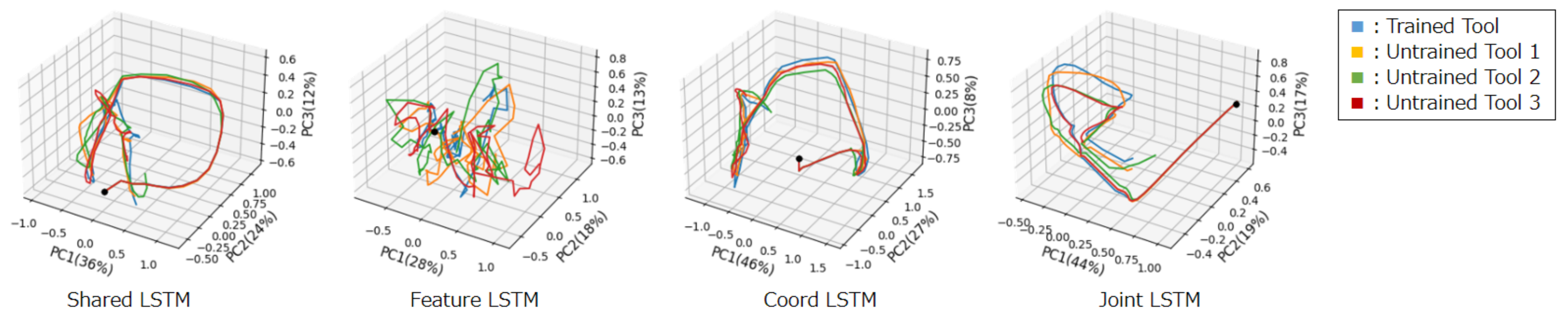}
  \caption{
    Comparison of transitioning internal states of layered LSTMs. The data were
    collected while predicting the dragging motion using a trained tool and
    three untrained tools. The states are compressed to three dimensions using
    principle component analysis, where the colored lines represent the transition
    route of each internal states. The black dot represents the initial state.
  }
  \label{fig:pca_comparison}
 \vspace{-3mm}
\end{figure*}

\begin{figure}
\centering
  \includegraphics[width=0.8\linewidth]{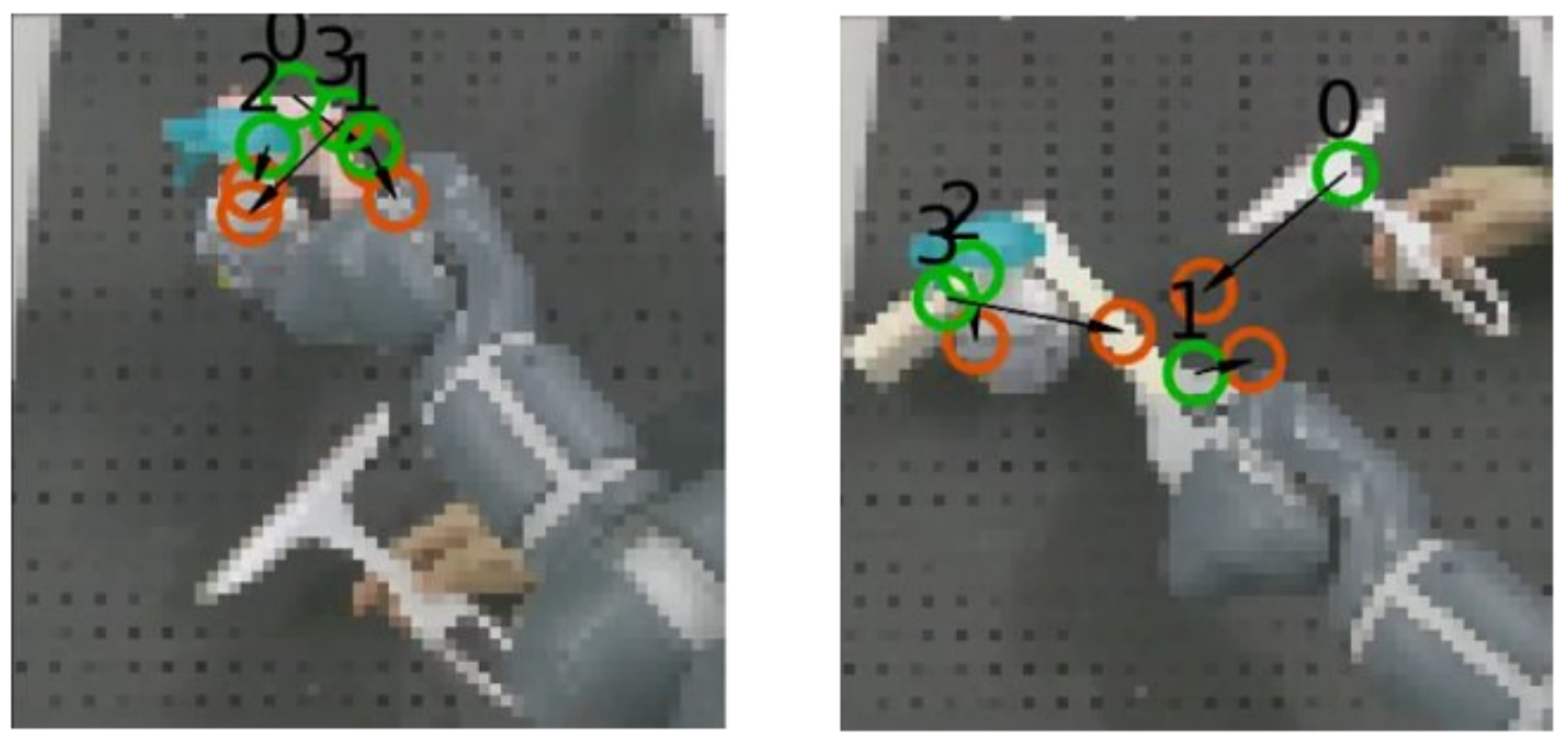}
  \caption{
     Sample results of predicted attention points when dragging with untrained
     tools. The end-effector attention (\#1) ignored the trained tool throughout
     the motion.
  }
  \label{fig:distractor_results}
 \vspace{-4mm}
\end{figure}

\subsubsection{Active + Single}
This model acquired attention for the necessary targets and showed active target
modification during motion. However, the role-based attention showed inconsistent
behaviors which tended to ignore the object after the dragging motion. Consequently,
the model lost track of the object during the picking motion and failed to
pick the object.

\subsubsection{Passive + Layered}
Although this model acquired appearance-based attention for the necessary targets,
the model was vulnerable to appearance changes during motion. This caused 
the attention points to fluctuate between different targets unnecessarily, producing
unstable robot motions.

\subsubsection{Passive + Single}
This model failed to acquire the necessary targets and showed vulnerability
to appearance changes.

Accordingly, the results suggests the following feature of each structure. The
``Active'' attention structure enabled the model to actively modify attention
targets and also be robust to appearance changes. The ``Layered'' LSTM
structure contributed to acquire role-based attentions that are coherent
across different tasks.

\subsection{Evaluation of tool-body assimilation}
Considering the fact that the proposed model perceived both the tool and gripper
as end-effectors (Fig. \ref{fig:normal_results}), we believe that the tool-body
assimilation had emerged. For further analysis, an additional experiment was
conducted to evaluate whether the model is capable of actively redefining the
body schema (c.f. Section \ref{sec:tool_use_learning}). The model was evaluated
by being presented with an untrained tool for performing dragging motions.

As Fig. \ref{fig:tool_results} shows, the trained model generalized to use all
the untrained tools and succeeded dragging the object. In addition, the previously
mentioned role-based end-effector attention directed to the edged parts of the
tools, where the robot actually hooked the object with. The attention on the
untrained tools retained throughout the dragging motions, including those that
had completely different appearances. Fig. \ref{fig:pca_comparison} compares
the transitioning internal states of the Layered LSTM while dragging using
trained and untrained tools. The internal states of Shared, Coord, and Joint
LSTMs transitioned similarly under all conditions, but slightly diverged after
grasping the tools. This indicates that the model generated different motions
for different tools, presumably based on the position of the predicted end-effector
attention. In contrast, the states of the Feature LSTM diverged drastically after
grabbing different tools; clear transitions are shown in video. This indicates
that the attention module sustainably directed attention for each tool as the
end-effector, with the knowledge that each tool is different. Since the model
is not provided with any information of untrained tools, the first direction of
the attention to the tools may have been an accident. However, the results
indicate that the  model actively modified the attention targets, since it
ignored the original tool after grabbing an untrained tool (Fig.
\ref{fig:distractor_results}). This suggests that the model stored the feature of
the end-effector attention inside the internal states and actively updated the
expected appearance of the tool. Such behavior of actively redefining
the end-effector (or body schema) is analogous to that of Japanese macaques in
Iriki's work \cite{iriki}, further suggesting the occurrence of tool-body
assimilation.
\section{CONCLUSION}
This paper proposed a novel real-time robot motion generation model, which
consists of an active top-down attention module and a layered LSTM motion predictor
module. Designating attention targets based on active feedback not only improved
the stability of conventional appearance-based attention, but also acquired
role-based attention that actively modifies attention targets according to the
situation. When trained on a tool-use task, such role-based attention perceived the
tool and the robot gripper as the same end-effector, which is analogous to the
biological tool-body assimilation. Thus, the introduction of active visual attention
extended the model's capability to perceive the environment flexibly and adaptively.

The proposed model is known to be capable of switching behaviors by control
via external inputs. This feature enabled the model to handle camera movements
during prediction in simulator environments, which we plan to apply to real-world
robot tasks in the future work. Further, the current limitation in tool-use tasks
is that generalizing to untrained tools is considered unstable because the initial
recognition depends on an accidental behavior. This issue is also planned to be
addressed in the future work.


\ifCLASSOPTIONcaptionsoff
  \newpage
\fi

\bibliographystyle{IEEEtran}
\bibliography{IEEEabrv,bibliography}

\end{document}